\documentclass[runningheads]{llncs}
\usepackage{multirow}
\usepackage{chngcntr} 
\usepackage{color}
\counterwithin{figure}{section}
\counterwithin{table}{section}
\usepackage{mwe} 
\usepackage{amsmath}
\usepackage{tabularx}
\usepackage{array}
\usepackage{amssymb}
\usepackage[capitalize]{cleveref}
\usepackage[T1]{fontenc}
\usepackage{latexsym}
\usepackage{amssymb}
\usepackage[capitalize]{cleveref}
\usepackage{multirow}
\usepackage{algpseudocode}
\usepackage{appendix}
\usepackage{float}
\usepackage{newfloat}
\usepackage{tabularx} 
\usepackage{ragged2e} 
\usepackage{dsfont}

\usepackage{marvosym}
\usepackage{graphicx}
\usepackage{color}

\begin{document}
\title{Class-Agnostic Region-of-Interest Matching in Document Images}
% \authorrunning{Anonymous Submission}
% Single author syntax
% \author{
    % Anonymous ICDAR Submission
    % email@example.com
% }
\author{Demin Zhang\inst{1}
\orcidID{0009-0006-0069-5190}\textsuperscript{*} \and Jiahao Lyu\inst{2, 3}\orcidID{0000-0003-2051-8045}\textsuperscript{*} 
\and
Zhijie Shen\inst{4}\orcidID{0009-0008-6135-7958}
\and
Yu Zhou\inst{1}\orcidID{0000-0003-4188-9953}\textsuperscript{\Letter}} 
\authorrunning{D. Zhang et al.}
\institute{
% Paper ID: 7}
VCIP \& TMCC \& DISSec, College of Computer Science \& College of Cryptology and Cyber Science, Nankai University, China
\\
\email{2312141@mail.nankai.edu.cn},
\email{yzhou@nankai.edu.cn}
\and
Institute of Information Engineering, Chinese Academy of Science, China \and School of Cyber Security, University of Chinese Academy of Science, China 
\email{lvjiahao@iie.ac.cn}
% \url{http://www.springer.com/gp/computer-science/lncs} 
\\
\and
Department of Computer Science and Technology, Tsinghua University, China 
\email{shenzj24@mails.tsinghua.edu.cn}
}

\maketitle                    % typeset the header of the contribution
\begin{abstract}
% The abstract should briefly summarize the contents of the paper in 150--250 words.
% 结合introduction重写

Document understanding and analysis have received a lot of attention due to their widespread application. However, existing document analysis solutions, such as document layout analysis and key information extraction, are only suitable for fixed category definitions and granularities, and cannot achieve flexible applications customized by users.
% In document analysis, key information extraction is receiving wide attention from researchers. It aims to extract vital information that the user requires. Visual element location is the first stage of key information extraction. However, the class definition is agnostic to document images from different sources.
Therefore, this paper defines a new task named ``Class-Agnostic Region-of-Interest Matching'' (``RoI-Matching'' for short), which aims to match the customized regions in a flexible, efficient, multi-granularity, and open-set manner.  The visual prompt of the 
% \abl{source} 
reference document and target document images are fed into our model, while the output is the corresponding bounding boxes in the target document images. 
% The whole pipeline is suggested to follow Open-set, Multi-granularity, and Customized. 
To meet the above requirements, we construct a benchmark RoI-Matching-Bench, which sets three levels of difficulties following real-world conditions, and propose the macro and micro metrics to evaluate. Furthermore, we also propose a new framework RoI-Matcher, which employs a siamese network to extract multi-level features both in the reference and target domains, and
cross-attention layers to integrate and align similar semantics in different domains.
% To solve this problem, we propose a simple Siamese network as our baseline to find the corresponding key areas of the visual prompt of source document images. Furthermore, we collect open-source document images and labels for benchmark construction at three granularities. 
% Experiments show that our method achieves available performance using a simple procedure. 
Experiments show that our method with a simple procedure is effective on RoI-Matching-Bench, and serves as the baseline for further research. The code is available at \url{https://github.com/pd162/RoI-Matching}.

\keywords{RoI-Matching  \and Document Analysis \and Benchmark.}
\end{abstract}

\let\thefootnote\relax
\footnotetext{*~Equal Contribution. \Letter~Corresponding Author.}

\section{Introduction}
% Introduction 写之前需要有逻辑
% 基本逻辑是：文档分析有应用价值 -> 目前的子任务不能做到高效感兴趣匹配 -> 感兴趣匹配与其它子任务的区别 -> 简要介绍我们做的方法和数据集 -> 总结三点贡献
% 然后画一张首图直接指出几个任务之间的差异

%首先介绍RoI-Matching的作用和意义

%Documents play a crucial role in storing and transmitting information, and how to understand their content and structure is a key issue. Consequently, some document analysis and understanding challenges like information retrieval, content extraction, digital archiving, and automated document parsing, have attracted increasing attention due to their broad applicability across various fields. To enhance document understanding, numerous downstream tasks have been developed to address different aspects of document understanding. For example, key information extraction aims to reveal the vital information in the document image, and document layout analysis is responsible for locating the important components of documents. These tasks contribute to various real-world applications, such as automating document processing in legal and financial affairs.
Document images play a vital role in storing and transmitting information, making content and structure understanding a key challenge.  As a result, tasks like processing~\cite{shu2025visualtextprocessingcomprehensive,zeng2024textctrl}, detection~\cite{cao2025devil,Shu_Perveiving,chen2021self}, recognition~\cite{qiao2020seed,qiao2021pimnet,yang2025ipad,zhang2025linguistics,shen2023divide}, spotting~\cite{lyu2025arbitrary,wang2022tpsnet,lyu2025textblockv2}, retrieval~\cite{li2025beyond,zeng2024focus}, and understanding~\cite{shen2025ldp,zeng2021beyond} have gained attention for their broad applicability.  To advance document understanding, various downstream tasks have been developed.  For example, key information extraction identifies vital content, while layout analysis locates important document components.  These tasks support real-world applications such as automating processes in legal and financial domains.
% For instance, in document understanding, key information extraction (KIE) aims to identify specific pieces of information from documents, such as dates, names, or addresses. Layout analysis focuses on understanding the structural organization of a document, including distinguishing between text, tables, and images. Template matching is used to identify predefined patterns in documents, often useful for processing standard forms. While these tasks have made significant progress, they often rely on fixed structures or well-defined templates and face challenges when dealing with more flexible and dynamic documents.
\begin{figure}[t]
\centering
\includegraphics[width=\textwidth]{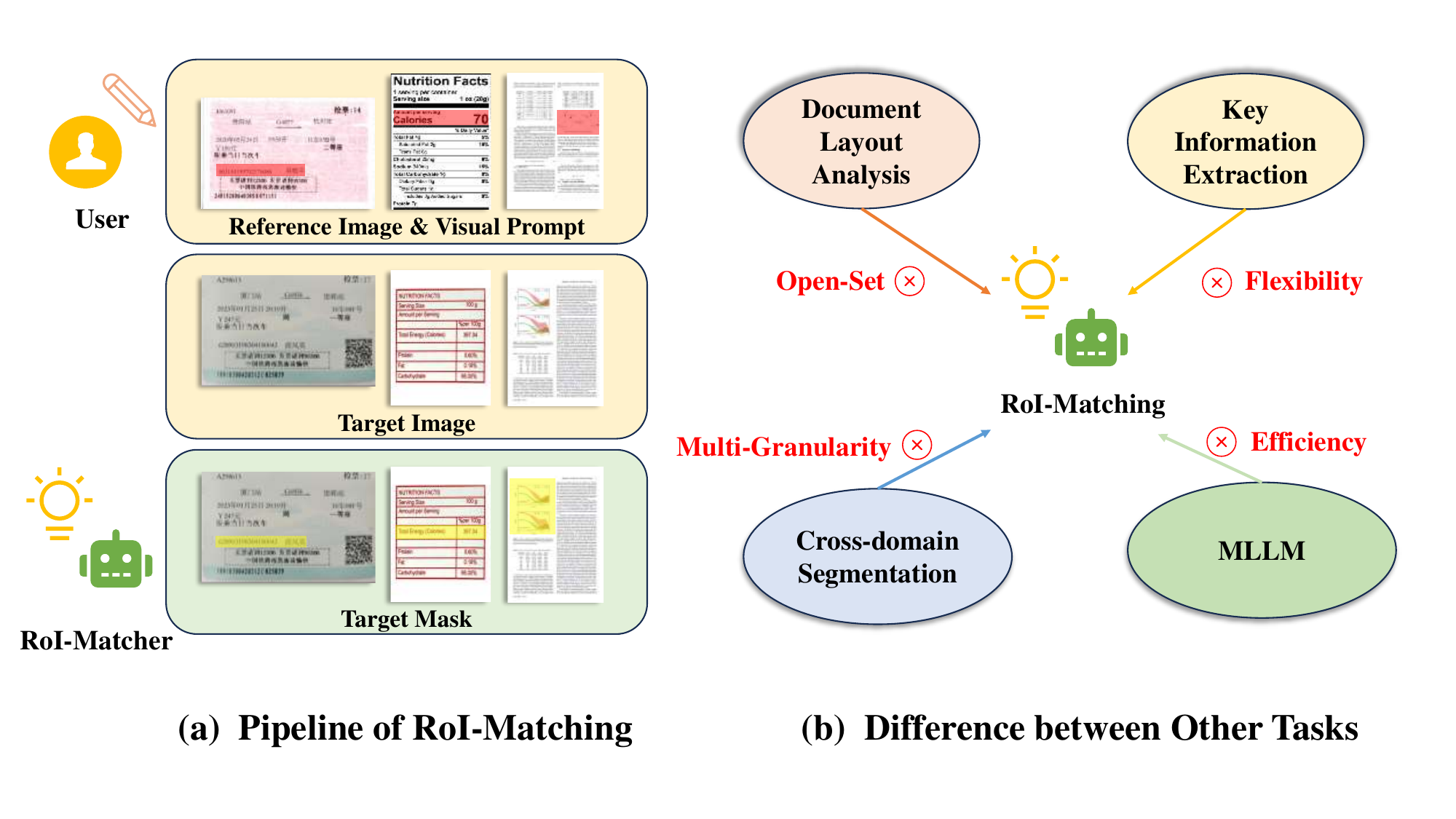}
    \caption{The task definition of RoI-Matching. (a) shows the pipeline of RoI-Matching. Given some document images by the user, RoI-Matcher responds to the corresponding mask, even if the user only provides visual prompt without class labels (``name'', ``number of calories'', and ``figure'' from left to right). (b) demonstrates the difference of RoI-Matching between other Tasks.}
    \label{fig:task}
    \vspace{-20pt}
\end{figure}

% 目前的问题

Despite significant advancements in downstream tasks, challenges remain in effectively matching user-customized regions of interest (RoIs) in documents, which is essential for practical applications in smart offices and ensuring user-friendly and flexible interactions. This difficulty arises because customized RoIs cannot be captured by a limited set of predefined concepts. Current document layout analysis methods are constrained by fixed categories, which hinder their generalization in open-world settings. As documents continue to diversify and evolve, developing more robust methods for efficient RoI matching becomes crucial, particularly in the presence of complex and unstructured content.

% 任务定义

In this paper, we propose a novel task, class-agnostic region-of-interest matching (RoI-Matching), aimed at efficiently matching user-defined regions of interest in document images. RoI-Matching is designed to match customized regions in a \textbf{flexible, efficient, multi-granular, and open-set} manner. \textbf{``Flexible''} refers to the diversity of user inputs, which are not restricted by language descriptions. To accommodate this flexibility, we introduce the concept of a visual prompt. Given the flexibility of the mask, we use a reference mask corresponding to the reference image to represent the customized visual prompt. \textbf{``Efficient''} indicates that the RoI-Matching is designed for high inference speed, ensuring its practical application. \textbf{``Multi-granular''} implies that the model must handle a range of scenarios, from word-level to line-level and paragraph-level instances. \textbf{``Open-set''} denotes the absence of class limitations in this task, ensuring its ability to generalize. The general inputs and outputs are illustrated in \cref{fig:task} (a). 

% 与其他任务的不同

RoI-Matching differs from other related tasks, as shown in \cref{fig:task}(b). First, document layout analysis \cite{zhong2019publaynet,yang2017learning,zhang2021vsr,shen2023divide} detects and segments different types of regions and analyzes their relationships within the document image. However, it is limited to fixed categories and cannot handle open-set scenarios. Key information extraction \cite{cao2023attention,hong2022bros,li2021structext} involves mining and interpreting multi-modal (semantic, layout, and visual) cues from visually rich document images to extract text information for specified categories. However, this process lacks flexibility, as it is only applicable to structured visually rich documents. Few-shot cross-domain segmentation \cite{lei2022cross,tavera2022pixel,huang2023restnet,herzog2024adapt} aims to learn a model that can perform segmentation on novel classes with only a few pixel-level annotated images. Although this is an open-set task, it is unsuitable for multi-granularity scenarios within document environments. In recent years, Multi-modal Large Language Model (MLLM) \cite{wang2024qwen2,chen2024internvl} solves many challenging problems of document images in an end-to-end manner. But the low inference speed of MLLMs fails to meet the efficiency requirements due to the limitations of auto-regressive decoding. Furthermore, studies have shown that MLLMs perform poorly in localization tasks. Thus, the introduction of RoI-Matching and the construction of the related dataset are meaningful for further enabling flexible and efficient understanding of document images.

% 本文做法

%To meet the above requirements, we construct a benchmark for RoI-Matching task and propose a simple yet effective baseline, RoI-Matcher. Specifically, we first curate a new dataset RoI-Matching-Bench, designed for the task of interest matching, enabling researchers to evaluate the performance of the RoI-Matching approach in real-world conditions. This dataset includes a variety of document types, ranging from standardized forms to more complex and unstructured documents, to assess the generalization capabilities. Furthermore, we introduce the RoI-Matcher, a simple yet efficient approach for this task in documents. In detail, we utilize a siamese network to automatically extract multi-granularity features of reference and target images in parallel. We then apply two cross-attention layers to project the regions of the reference image into the target domain. Additional design considerations mitigate the impact of overlapping responses, which are common in document images. In conclusion, our contributions in this paper are threefold:
To meet the above requirements, we construct a benchmark for the RoI-Matching task and propose a simple yet effective baseline, RoI-Matcher. Specifically, we curate a new dataset, RoI-Matching-Bench, designed for region-of-interest matching in documents, enabling performance evaluation under real-world conditions. The dataset covers diverse document types, from standardized forms to complex unstructured layouts, to assess generalization.We also introduce RoI-Matcher, a simple yet efficient method that uses a Siamese network to extract multi-granularity features from reference and target images in parallel. Two cross-attention layers are used to project reference regions into the target domain, while additional designs mitigate overlapping response issues common in document images. Our main contributions are threefold:

\begin{itemize}

\item 
We define a new task, RoI-Matching, to match the customized regions in a flexible, efficient, multi-granular, and open-set manner. 

\item  
We also construct the corresponding RoI-Matching-Bench with three levels of difficulties, which allows researchers to evaluate the performance of RoI-Matching approach under real-world conditions. Two evaluation metrics are also introduced in the macro and micro perspectives.

\item
We propose a new framework RoI-Matcher, which employs a siamese network to extract multi-level features both in the reference and target domains, and cross-attention layers to integrate and align similar semantics in different domains. Some additional designs also mitigate the impact of overlapping responses and improve performances.
\end{itemize}

\section{Related Works}

\subsection{Document Layout Analysis}
%先介绍方法的发展

%Document layout analysis (DLA) methods can be divided into uni-modal and multi-modal methods. Uni-modal layout analysis primarily focuses on leveraging visual features to analyze document layouts. Several methods are derived from object detection and instance segmentation techniques for detecting and segmenting document regions. For instance, PubLayNet \cite{zhong2019publaynet} applies Faster R-CNN \cite{ren2016faster} and Mask-RCNN \cite{he2017mask} for document layout analysis. TransDLANet \cite{cheng2023m6doc} uses three parameter-shared multi-layer perceptions on top of ISTR, while SwinDocSegmenter \cite{banerjee2023swindocsegmenter} merges both high-level and low-level features of document images to initialize the query in DINO \cite{zhangdino}. SelfDocSeg \cite{maity2023selfdocseg} generates pseudo-layouts to pretrain the image encoder and then fine-tunes it on layout analysis datasets, using the BYOL \cite{grill2020bootstrap} framework.
Document layout analysis (DLA) methods fall into uni-modal and multi-modal categories. Uni-modal approaches focus mainly on visual features to analyze layouts, often adapting object detection and instance segmentation techniques. For example, PubLayNet \cite{zhong2019publaynet} employs Faster R-CNN \cite{ren2016faster} and Mask R-CNN \cite{he2017mask} for layout detection. TransDLANet \cite{cheng2023m6doc} utilizes three parameter-shared MLPs atop ISTR, while SwinDocSegmenter \cite{banerjee2023swindocsegmenter} combines high- and low-level image features to initialize queries in DINO \cite{zhangdino}. SelfDocSeg \cite{maity2023selfdocseg} pretrains its image encoder with pseudo-layouts using the BYOL \cite{grill2020bootstrap} framework before fine-tuning on layout datasets.

While these approaches improve model performance, they remain limited in fully leveraging document semantics. Recently, multi-modal methods have been introduced to address complex DLA challenges. For example, MFCN \cite{yang2017learning} employs skip-gram to capture sentence-level textual features and integrates them with visual features in the decoder. VSR \cite{zhang2021vsr} builds a full text embedding map combining Char-grid, word-grid, and sentence-grid levels, using dual backbones to extract and fuse visual and textual features via a multi-scale-adaptive-aggregation module. Notable models such as LayoutLM \cite{xu2020layoutlm,xu2021layoutlmv2,huang2022layoutlmv3}, BEiT \cite{baobeit}, DiT \cite{li2022dit}, UDoc \cite{gu2021unidoc}, and StrucText \cite{li2021structext,yu2023structextv2} adopt a pre-training then fine-tuning approach. Extensive pre-training on large document datasets enables these models to excel in downstream tasks like image classification, layout analysis, and information extraction.

\subsection{Cross-domain few-shot Segmentation}

% Cross-domain Few-Shot Semantic Segmentation (CD-FSS) is a semantic segmentation task that combines cross-domain adaptation and few-shot learning.  It aims to segment novel classes in a test (query) image based on a few labeled (support) images.

%Cross-domain few-shot semantic segmentation (CD-FSS) aims to segment unseen classes in query images with only a limited number of annotated samples. PATNet \cite{lei2022cross} first proposes this task and designs a feature transformation module to convert domain-specific features into domain-agnostic features. RestNet \cite{huang2023restnet} employs a lightweight attention module to enhance pre-transformation features and merge post-transformation features through residual connections to maintain the key information in the original domain. RD \cite{tavera2022pixel} employs a memory bank to restore the meta-knowledge of the source domain to augment the target domain data. APSeg \cite{he2024apseg} first adapt SAM into CD-FSS task with the introduction of dual prototype
%anchor transformation (DPAT) module and a meta prompt generator (MPG) module efficiently. ABCDFSS \cite{herzog2024adapt} introduces a novel consistency-based contrastive learning scheme, which can estimate the parameters of our attached layers without overfitting to the support set and provide domain-shift robust prediction masks. Due to the similar types of inputs and outputs, we regard these methods as comparators to prove our method.
Cross-domain few-shot semantic segmentation (CD-FSS) targets segmenting unseen classes in query images using only a few annotated samples. PATNet \cite{lei2022cross} pioneered this task by designing a feature transformation module to convert domain-specific features into domain-agnostic ones. RestNet \cite{huang2023restnet} enhances pre-transformation features with a lightweight attention module and merges post-transformation features via residual connections to preserve key domain information. RD \cite{tavera2022pixel} uses a memory bank to restore source domain meta-knowledge and augment target data. APSeg \cite{he2024apseg} adapts SAM to CD-FSS with a dual prototype anchor transformation (DPAT) module and a meta prompt generator (MPG) for efficient learning. ABCDFSS \cite{herzog2024adapt} introduces a consistency-based contrastive learning scheme that estimates attached layer parameters without overfitting and yields domain-shift robust masks. Given their similar input-output settings, we compare our method against these approaches.

% Shuo Lei et al\cite{lei2022cross}. established a new benchmark for CD-FSS to evaluate the cross-domain generalization ability of few-shot segmentation models under different domain offsets, which include four different domain datasets FSS-1000, Deepglobe, ISIC2018, Chest X-ray.  Experimental results show that the performance of existing methods decreases significantly when the domain difference is large.
% The author also proposes a new model, PATNet, which solves the CD-FSS problem by transforming domain-specific features into domain-independent features and achieves superior performance on the CD-FSS benchmark.

\subsection{Key Information Extraction}

Currently, methods for Key Information Extraction (KIE) can be divided into two main types: OCR-dependent and OCR-free models. OCR-dependent models use optical character recognition (OCR) to extract textual information. Early KIE methods \cite{wang2021layoutreader,zhang2023reading} construct layout-aware or graph-based representations for KIE tasks with sequence labeling of OCR inputs. Nevertheless, these methods rely on text with a proper reading order or need additional modules for OCR serialization, which is not always feasible where the document layout might be intricate or unordered in real-world scenarios. To tackle the serialization problem, certain methods \cite{hong2022bros,xu2021layoutxlm} make use of extra detection or linking modules to model the intricate connections between text blocks or tokens. Additionally, methods based on generation \cite{cao2022query,cao2023attention,tang2023unifying} have been put forward to reduce the effort of post-processing and the need for task-specific link designs. Another category of OCR-free methods \cite{lee2023pix2struct,kil2023prestu} offers an alternative, which either utilize OCR-aware pre-training or incorporates OCR modules within an end-to-end framework. Donut\cite{kim2022ocr}, Omniparser\cite{wan2024omniparser} and other methods adopt a text reading pretraining objective and generate structured outputs consisting of text and entity token.

\section{Benchmark}
\subsection{Statistics and Visualizations}
% 数据来源和收集流程 向其他数据集论文学习出一点统计学意义上的图表明广泛分布 三个子集挑一些样例

% We divided the tasks into three difficulty levels and collected appropriate images from multiple open-source datasets based on these three difficulty levels to construct three benchmarks.
\begin{figure}[t]
\centering
\includegraphics[width=\textwidth]{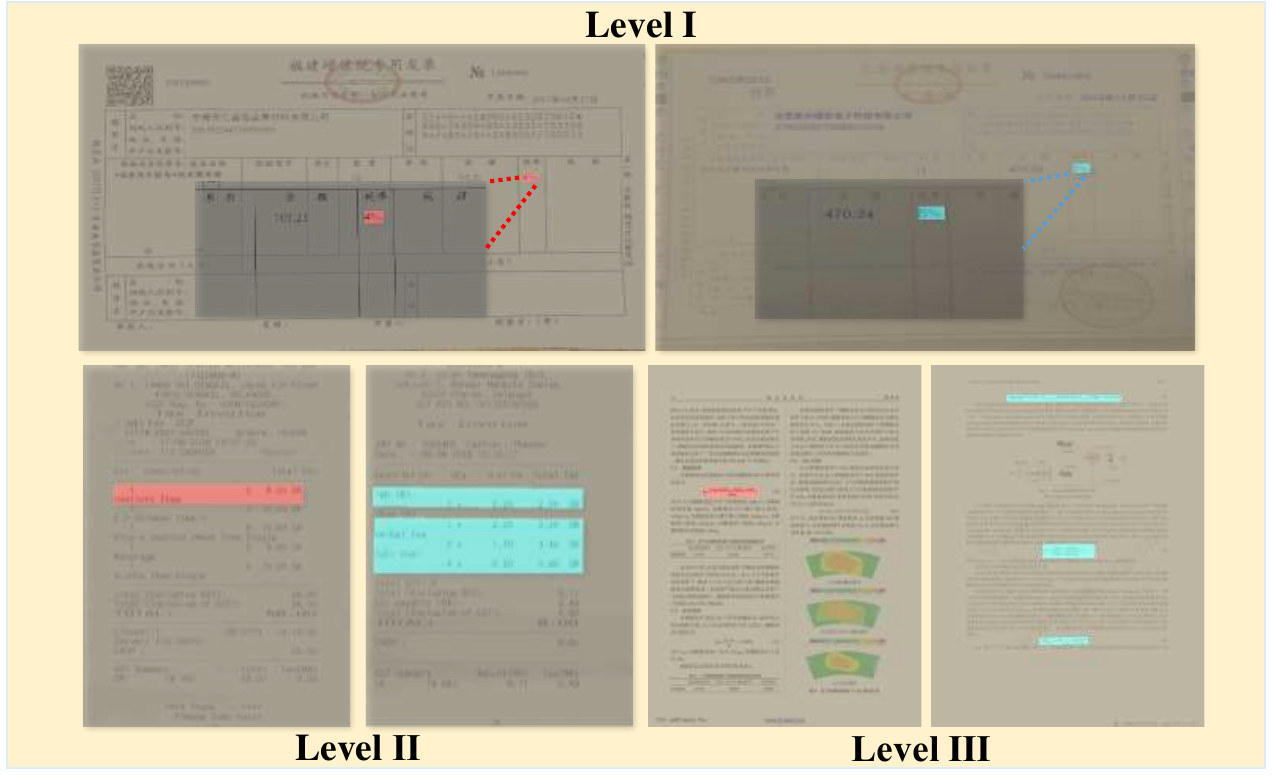}
    \caption{The visualizations of RoI-Matching-bench. The reference image (left) and the target image (right) are image pairs. Hint: these visual prompts indicate semantic labels, which are ``Tax Ratio'', ``Items in Receipt'', ``Equation''.}
    \label{fig:dataset}
    \vspace{-10pt}
\end{figure}

%To build the benchmark RoI-Matching-Bench for the RoI-Matching task, we utilize six existing document datasets covering key information extraction, document layout analysis and more. Details on dataset distribution and sources are presented in \cref{tab:dataset}. The task is divided into three difficulty levels—Level \uppercase\expandafter{\romannumeral1}, Level \uppercase\expandafter{\romannumeral2}, and Level \uppercase\expandafter{\romannumeral3}—based on matching granularity and complexity. For Levels \uppercase\expandafter{\romannumeral1} and \uppercase\expandafter{\romannumeral3}, masks are generated directly from open-source box annotations. For Level \uppercase\expandafter{\romannumeral2}, a semantic category schema is first defined, followed by manual annotation from trained annotators. All annotations undergo rigorous visual verification to ensure accuracy and consistency. We group semantically and visually similar images into categories, pair them, and randomly sample to balance distribution and avoid outliers. We select categories with distinctive features (e.g., bank card numbers, forms) and sample corresponding images, excluding content without significant visual features, such as article paragraphs. For datasets with predefined splits, we use the original partitions, for datasets lacking validation sets, we assign a stratified random subset of the training set as the validation set. Definitions of the three levels are provided below.
To build the RoI-Matching-Bench benchmark for the RoI-Matching task, we leverage six existing document datasets spanning key information extraction, document layout analysis, and more. Dataset details and sources are listed in \cref{tab:dataset}. The task is divided into three difficulty levels—Level I, II, and III—based on matching granularity and complexity. For Levels I and III, masks are generated directly from open-source box annotations. For Level II, we first define a semantic category schema, followed by manual annotation from trained annotators. All annotations undergo strict visual verification to ensure accuracy and consistency. We group semantically and visually similar images into categories, pair them, and randomly sample to balance the distribution and avoid outliers. Categories with distinctive features (e.g., bank card numbers, forms) are selected along with corresponding images, while content lacking significant visual features, such as article paragraphs, is excluded. For datasets with predefined splits, we use the original partitions; for those without validation sets, we create stratified random subsets from the training data as validation sets. Definitions of the three levels are provided below.

\begin{table}[t]
\centering
\caption{Derivation, scenarios, and statistics on the benchmark of RoI-Matching.}
\label{tab:dataset}
\renewcommand{\arraystretch}{1.0}
\begin{tabular}{ccccccc
}
\hline
\multicolumn{1}{c}{\multirow{2}{*}{Difficulties}} & \multicolumn{1}{c}{\multirow{2}{*}{Derivation}} & \multirow{2}{*}{Images} & \multicolumn{3}{c}{Pairs} & \multicolumn{1}{c}{\multirow{2}{*}{Examples of Scenario}} \\ \cline{4-6}
\multicolumn{1}{c}{}                           & \multicolumn{1}{c}{}                            &                         & Train    & Val   & Test   & \multicolumn{1}{c}{}                           \\ \hline
Level \text{\uppercase\expandafter{\romannumeral1}}                               & SVRD  \cite{yu2023icdar}                                          & 660                     & 107k     & 253   & 1260     & Invoice, ID Card, Ticket.                      \\
Level \text{\uppercase\expandafter{\romannumeral2}}                                  & POIE \cite{kuang2023visual}, SROIE \cite{wang2021towards}                                    & 233                     & 13k      & 851   & 913& Receipt, Nutrition Fact.                      \\
Level \text{\uppercase\expandafter{\romannumeral3}}                                  & D4LA \cite{da2023vision}, M6Doc \cite{cheng2023m6doc}, etc.                           & 4312                    & 33k      & 809   & 673    & Paper, Advertisement.                         \\ \hline
\end{tabular}
\vspace{-10pt}
\end{table}

\textbf{Level \text{\uppercase\expandafter{\romannumeral1}}}. As the simplest and most common case, we define Level \uppercase\expandafter{\romannumeral1} as Reference-Target pairs with similar visual elements and positions. This level of matching is commonly found in scenarios like standardized forms, including invoices, ID cards, tickets, etc. To construct the dataset following the above definition, we regard ICDAR2023-SVRD \cite{yu2023icdar} as baseline benchmark and automatically create an organized benchmark, due to its visually similar images and detailed annotations.

\textbf{Level \text{\uppercase\expandafter{\romannumeral2}}}. Compared to Level \text{\uppercase\expandafter{\romannumeral1}}, Level \text{\uppercase\expandafter{\romannumeral2}} includes reference-target pairs that are visually similar but differ in position. We use two datasets from the field of key information extraction (KIE) to construct this sub-benchmark. To ensure consistency with other sub-benchmarks, we manually sample and annotate these data.

\textbf{Level \text{\uppercase\expandafter{\romannumeral3}}}. As the most challenging sub-task, {Level \text{\uppercase\expandafter{\romannumeral3}}} consists of the samples which are significantly different in both visual appearance and position. These samples are commonly found in advertisements and academic papers. We use three datasets from the field of document layout analysis and select classes that contain visual elements with distinct features and similar semantics, such as equations, figures, and titles of academic papers.

\begin{figure}[t]
\centering
\includegraphics[width=\textwidth]{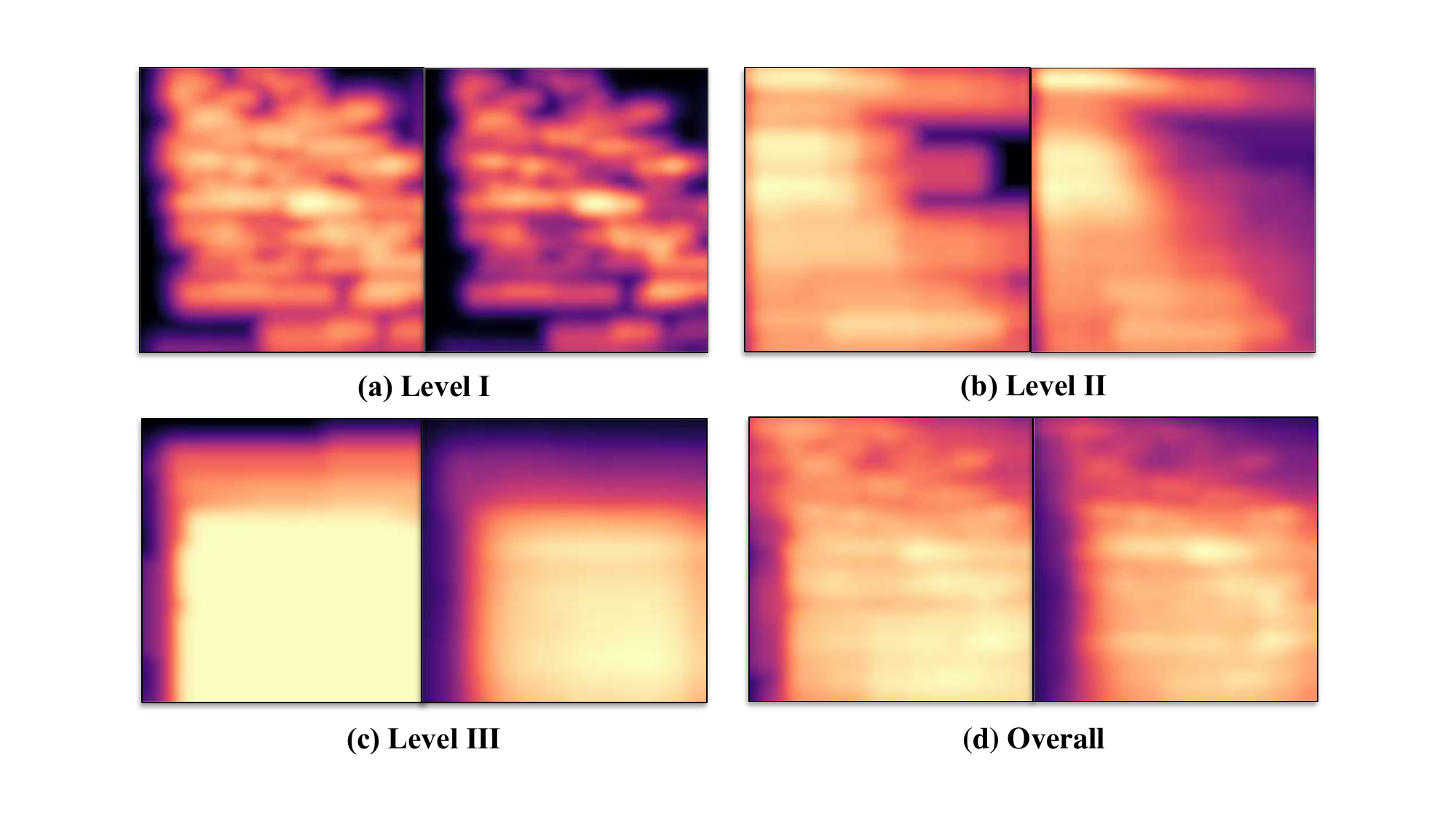}
    \caption{The position distribution of RoI-Matching benchmark. There are two images in each group, the left is the visualization of the reference images, and the right is the target images.}
    \label{fig:distribution}
    \vspace{-15pt}
\end{figure}

Furthermore, we gather statistics of RoI-Matching-Bench and visualize the position distribution in \cref{fig:dataset} and \cref{fig:distribution}, which demonstrate the construction of our benchmark across multi-grained data, satisfying the generalized rule.

\subsection{Evaluation Protocols}
% macro and micro
We report two metrics: one from the pixel-wise protocol for common semantic segmentation and another from the instance-wise protocol for scene text detection.

\noindent\textbf{Mean Interaction-of-Union (mIoU)}. Inspired by \cite{long2015fully}, we use mIoU to evaluate the effectiveness of our method from the pixel-level perspective. The formulation of mIoU is shown as \cref{eq:miou}. Let $p_{ij}$ represent the number of pixels of class $i$ predicted to belong to class $j$, where there are $k$ different classes. $k=1$ in this paper due to the setting of class-agnostic.

\begin{equation}
  mIoU = \frac{1}{k+1}\sum_{i=0}^{k}\frac{p_{ii}}{\sum_{j=0}^k p_{ij} + \sum_{j=0}^k p_{ji} - p_{ii}}.
\label{eq:miou}
\end{equation}

\noindent\textbf{F-Measure}.  F-measure is widely used to evaluate binary classifiers in machine learning. In our evaluation process, we first convert the final binary mask into instance-level prediction through connected component analysis. Then, we calculate IoU score between the ground truth and the prediction. We consider the samples with $IoU >0.5$ as correctly predicted. Following standard practices in scene text detection, we use precision, recall, and F-measure to evaluate the RoI-Matching results.

\begin{equation}
\text{F-Measure} = \frac{2 \cdot \text{Precision} \cdot \text{Recall}}{\text{Precision} + \text{Recall}}.
    \label{eq:f}
\end{equation}

\section{Method}

\begin{figure}[t]
\centering
\includegraphics[width=\textwidth]{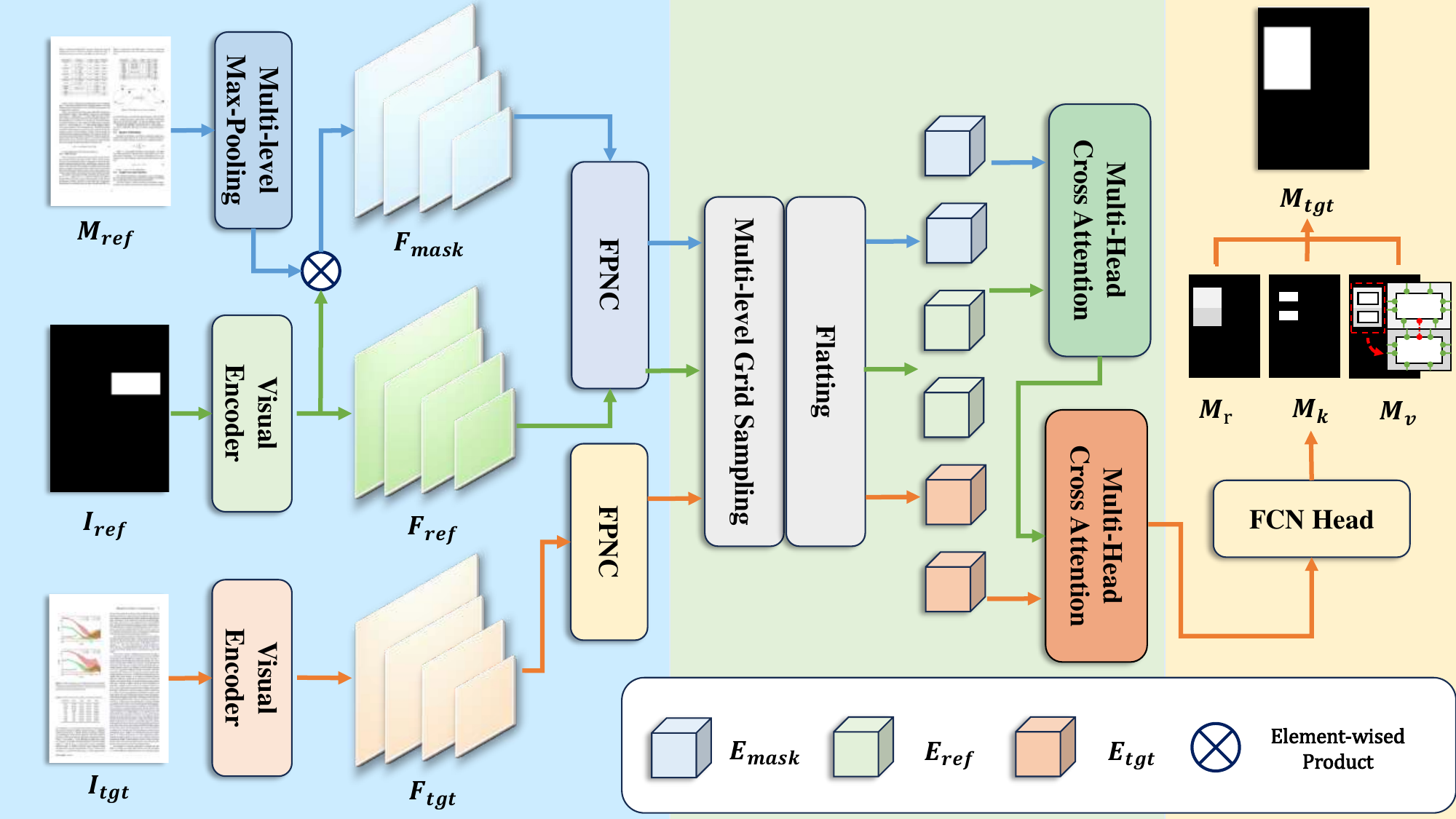}
    \caption{The architecture of RoI-Matcher. The background colors represent three stages of our model, blue for Visual Perception, green for Cross-granularity Attention, and yellow for Segmentation Head, respectively. The colors of arrows indicate flows of different inputs, blue for the reference mask, green for the reference image, and red for the target image.}
    \label{fig:model}
    \vspace{-10pt}
\end{figure}

% 按照我那个PPT写 我完成这一部分
In this section, we introduce the class-agnostic region-of-interest matching task first. Then an effective baseline for our benchmark is described in detail. The baseline consists of three main components: Visual Perception, Cross-granularity Attention, and Segmentation Head. Furthermore, we introduce task-aware label generation and optimization.

\subsection{Task Definition}

In this section, we first propose the class-agnostic region-of-interest matching task in document scenarios. This task can be formalized as follows. The prior are reference document image $I_{ref} \in \mathbb{R}^{H_{ref} \times W_{ref} \times 3}$ and user-customized visual prompt, representing a binary mask $M_{ref} \in \mathbb{R}^{H \times W}$. Because a mask can represent any form of visual prompt.  
% \abl{(mask should be a bounding box with two points) for $I_{ref}$}. 
During the inference stage, given the target document image $I_{tgt} \in \mathbb{R}^{H_{tgt} \times W_{tgt} \times 3}$, 
% \abl{(Ref vs tgt)}, 
the model $\mathcal{M}$ aims to find the region of interest, where a binary mask $M_{tgt} \in \mathbb{R}^{H_{tgt} \times W_{tgt}}$ for $I_{tgt}$ represents the final result. Note that there is no category information during the whole pipeline. Therefore, we define this task as the class-agnostic region-of-interest matching.

\subsection{Visual Perception}

Visual Perception aims to extract multi-level features $F_{ref}$ and $F_{tgt}$ from reference image $I_{ref}$ and target image $I_{tgt}$. The extraction is formulated by \cref{eq:f_k}, where $k$ is the index of $\{ref, tgt\}$.

\begin{equation}
   F_{k} = VE(I_{k}) \in \{\mathbb{R}^{\frac{W_{k}}{4}\times \frac{W_{k}}{4} \times C},
   \mathbb{R}^{\frac{W_k}{8}\times \frac{W_k}{8} \times 2C},
   \mathbb{R}^{\frac{W_k}{16}\times \frac{W_k}{16} \times 4C},
   \mathbb{R}^{\frac{W_k}{32}\times \frac{W_k}{32} \times 8C}
   \}.
\label{eq:f_k}
\end{equation}

Moreover, the reference mask provided by user $M_{ref}$ interacts initially with $F_{ref}$ level by level, as conducted by \cref{eq:f_mask}. 
$MP(x, y)$ indicates the Max Pooling operator, where $x$ is feature map and $y$ is the sampling ratio. $\otimes$ is the element-wise dot production. $i$ represents $i$-th layer.

\begin{equation}
   F_{mask}^{i} = F_{ref}^{i} \otimes MP(M_{ref}^{i}, 2^i).
\label{eq:f_mask}
\end{equation}

Then four feature maps 
% \abl{why 3 maps? equ.3 has 4 maps} 
are aggregated using the efficient FPNC \cite{liao2020real} respectively.
FPNC is a kind of FPN variant and more efficient than classical FPN. This process is formulated by \cref{eq:f_fpn}, where $j$ is the index of $\{ref, tgt, mask\}$. Note that $\mathcal{F}_{mask}$ and $\mathcal{F}_{ref}$ is generated by weight-shared FPNC, because these two feature maps are homologous.

\begin{equation}
   \mathcal{F}_{j} = FPNC(F_{j}).
\label{eq:f_fpn}
\end{equation}

\subsection{Cross-granularity Attention}
After extracting from the reference and target document image, the key problem is how to transfer from the reference visual prompt to the target domain. Considering the high-resolution of the document image, we first sample the multi-level features into smaller feature maps to alleviate the computational burden. Then the sampled features are flattened into different embeddings, termed as $E_{mask}, E_{ref}, E_{tgt}$. The grid sampling and flattening process is written as \cref{eq:gs}.  

\begin{equation}
   E_{j} = Flatten(GS(\mathcal{F
   }_{j})) \in \mathbb{R}^{\frac{H_k W_k}{256} \times 4C}
\label{eq:gs}
\end{equation}

Extracted by multi-level feature maps, $E_{mask}, E_{ref}, E_{tgt}$ contains rich vision information from different granularities. The interaction between different granularities and domains is realized by Multi-Head Cross Attention \cite{vaswani2017attention}. The detailed process is formulated as \cref{eq:mhca1} and \cref{eq:mhca2}. For the first cross attention, we regard the $E_{mask}$ as the query and $E_{ref}$ as the key and value. Two cross-attention operations ensure transfering the visual prompt from the reference domain to the target domain, preventing information loss. 

\begin{equation}
   E_{r} = MHCA(Q=E_{mask};K=E_{ref}; V=E_{ref}),
\label{eq:mhca1}
\end{equation}
\begin{equation}
   E_{total} = MHCA(Q=E_{tgt};K=E_{r}; V=E_{r}).
\label{eq:mhca2}
\end{equation}

\subsection{Segmentation Head}

After interacting features from different domains and granularities, we construct the Segmentation Head to generate the final mask $\hat{M}_{tgt}$. Considering the overlapped regions (as shown in $M_r$ in \cref{fig:model}) in document images, we introduce the subtle mask representation inspired by PAN \cite{wang2019efficient}.
Compared with a single binary mask, we use the 6-channel mask to represent $\hat{M}_{tgt}$. Naively, we adopt Fully Convolution Network (FCN) \cite{long2015fully} to predict 6 channels. The first channel is regions $M_r$, and the second is kernel $M_k$, where its label generation will be described in Section 4.5. The last four channels indicate similarity vectors, which guide the pixels in the text area to the corresponding kernel. The detailed post-processing follows PAN \cite{wang2019efficient}, where the $M_k$ decides the number of regions, then $M_v$ assists in refining the region more precisely.

\begin{equation}
   {M'}_{tgt} = [M_r, M_k, M_v] = UpSampling(FCN(E_{total})) \in \mathbb{R}^{6 \times H_t \times W_t} .
\label{eq:seg}
\end{equation}

\subsection{Label Generation}

Due to the dense information in document images, the regions of interest could be overlapped. Therefore, we need to use the strategy during the label generation stage. The label generation for the target mask $M_{k}$ is inspired by PSENet \cite{wang2019shape}. Specifically, given a target image with $N_r$ regions of interest, each region can be described by a set of segments, where $n$ is the number of vertexes. In this task, user-customized visual prompt is represented by a binary mask, which is regarded as an irregular polygon. So $n$ is not fixed.

\begin{equation}
   G = \{S_k\}_{k=1}^n.
\label{eq:G}
\end{equation}

The shrink offset $D$ of shrinking is computed from the perimeter $L$ and
area $A$ of the original polygon:
where $r$ is the shrink ratio, set to 0.4 empirically. The final label $M_{k}$ is generated by the original label $G$ and offset $D$, using the Vatti clipping algorithm \cite{vatti1992generic}.

\begin{equation}
   D = \frac{A(1-r^2)}{L}.
\label{eq:vatti}
\end{equation}

\subsection{Optimization}
Followed by PAN \cite{wang2019efficient}, our optimization target can be formulated as:

\begin{equation}
   \mathcal{L} = \mathcal{L}_{region}  + \alpha \mathcal{L}_{kernel} +\beta( \mathcal{L}_{agg} + \mathcal{L}_{dis}),
\label{eq:loss}
\end{equation}
where $\mathcal{L}_{region}$ is the loss of regions and $\mathcal{L}_{kernel}$ is the loss of kernels.  $\alpha$ and $\beta$ are the balanced factors, we set them to 0.5 and 0.25 respectively.

Considering segmentation task, we adopt dice loss \cite{milletari2016v} to supervise $\hat{M}_{r}$ and $\hat{M}_{k}$. $\mathcal{L}_{region}$ and $\mathcal{L}_{kernel}$ can be written as \cref{eq:loss_text} and \cref{eq:loss_kernel}. We also use  Online Hard Example Mining (OHEM) \cite{shrivastava2016training} to solve the problem of imbalanced foreground and background pixels.

\begin{equation}
   \mathcal{L}_{region} = 1 - \frac{2\sum_{i, j}\hat{M}_{r}(i,j)M_{r}(i,j)}{\sum_{i, j}\hat{M}_{r}(i,j)^2 + \sum_{i, j}{M}_{r}(i,j)^2},
\label{eq:loss_text}
\end{equation}

\begin{equation}
  \mathcal{L}_{kernel} = 1 - \frac{2\sum_{i, j}\hat{M}_{k}(i,j)M_{k}(i,j)}{\sum_{i, j}\hat{M}_{k}(i,j)^2 + \sum_{i, j}{M}_{k}(i,j)^2}.
\label{eq:loss_kernel}
\end{equation}

While calculating $\mathcal{L}_{agg}$, we regard the pixels in $M_r$ but not in $M_k$ as valid pixels, illustrated by $M_v$ in \cref{fig:model}. Given $N_{tgt}$ regions of interest in $I_{tgt}$, the valid pixel in i-th region can be formulated as $T_i$ and kernel as $K_i$. Then we use \cref{eq:loss_agg} to supervise the overlapped pixels aggregating the proper region, as shown green dashed line of \cref{fig:model}.

\begin{equation}
  \mathcal{L}_{agg} = \frac{1}{N_{tgt}}\sum_{i=1}^{N_{tgt}} \frac{1}{T_i}\sum_{p \in T_i} ln(1 + D(p, K_i)),
\label{eq:loss_agg}
\end{equation}

\begin{equation}
  \mathcal{L}_{dis} = \frac{1}{N_{tgt}(N_{tgt} - 1)}\sum_{i=1}^{N_{tgt}}\sum_{j=1, j\neq i}^{N_{tgt}} ln(1 + D(K_i, K_j)),
\label{eq:loss_dis}
\end{equation}
where $p$ is valid pixel of $i$-th region of interest. $D(\cdot)$ is the distance function. 
In addition, the cluster centers need to keep discrimination. Therefore, the kernels of different text instances should maintain enough distance as shown red dashline of \cref{fig:model}.

\section{Experiments}

We conduct extensive experiments to analyze the effectiveness of our proposed RoI-Matcher. In this section, we first introduce the implementation details, which are described in Sec. 5.1. Then we compare our RoI-Matcher with other CD-DSS methods due to their similar inputs and outputs. Finally,  the 
ablation studies are shown in Sec. 5.3 to analyze the effectiveness and efficiency of our proposed modules.

\begin{table}[t]
\centering
\caption{Comparison results of RoI-Matcher. $\dagger$ means only samples that fit the output format participate in the calculation during the evaluation.}
\setlength{\tabcolsep}{1pt}
\label{tab:comparison}
\renewcommand{\arraystretch}{1.3}
\resizebox{\linewidth}{!}{
\begin{tabular}{lcccccccc}
\hline
\multicolumn{1}{c}{\multirow{2}{*}{Methods}} & \multicolumn{2}{c}{Level I} & \multicolumn{2}{c}{Level II} & \multicolumn{2}{c}{Level III} & \multicolumn{2}{c}{Total} \\ \cline{2-9} 
\multicolumn{1}{c}{}                         & mIoU (\%)       & F (\%)     & mIoU (\%)      & F (\%)      & mIoU (\%)       & F (\%)      & mIoU (\%)     & F (\%)    \\ \hline
\multicolumn{9}{c}{Cross-Domain Few-Shot Segmentation} \\ \hline
PATNet \cite{lei2022cross}                                      &        65.1    &   46.7             &       60.9     &         40.1        &       48.6      &     15.9            &     56.8
         &     27.5         \\ 
RestNet \cite{huang2023restnet}                                      &     73.2          &      63.6           &      57.8      &       31.7          &    48.7         &        15.1         &        57.9   &     29.8          \\
DMTNet \cite{duong2023dmt}                                      &        70.3    &   58.2             &       61.9     &         42.7       &       50.1      &     20.4            &     59.5
         &     34.5         \\ 
ABCD-FSS    \cite{herzog2024adapt}                                   &        40.3    &   3.8             &         38.4   &         16.8       &       52.0      &     39.8            &    43.5
         &     9.2         \\ \hline
\multicolumn{9}{c}{Multimodal Large Language Model} \\ \hline
Qwen2-VL-7B \cite{wang2024qwen2} $\dagger$                                          &     67.4       &   65.2            &     63.2       &    62.3            &    48.7         &     33.8           &    57.3       &   55.3            \\ 
 InternVL2.5-8B \cite{chen2024internvl} $\dagger$                                   &     67.5       &   64.8            &     62.7       &    60.5            &   46.0         &     25.9         &    53.2     &   50.5            \\ 
\hline
\multicolumn{9}{c}{Class-Agnostic RoI-Matching} \\ \hline
RoI-Matcher (Ours)                           &      92.1      &   92.2             &   88.5         &    84.2            &   85.0          &       72.3          &   87.3        &      83.9         \\ \hline
\end{tabular}
}
% \vspace{-10pt}
\end{table}

\subsection{Implementation Details}

%We implement our RoI-Matcher with the native PyTorch library \cite{paszke2019pytorch}. The training and testing images are resized to 640 $\times$ 640, and data augmentation strategies including RandomFlip, RandomRotate, and RandomBrightnessContrast are applied. AdamW is used as an optimizer with a weight decay of 1e-4. The learning rate is initialized at 1e-4 and adjusted by step schedule. All experiments are conducted on NVIDIA RTX 2080 Ti GPUs, except for the MLLM inference experiment, which we performed on the NVIDIA RTX A6000.

We implement our RoI-Matcher using the native PyTorch library \cite{paszke2019pytorch}. Training and testing images are resized to 640×640, with data augmentations including RandomFlip, RandomRotate, and RandomBrightnessContrast. We use AdamW optimizer with a weight decay of 1e-4 and an initial learning rate of 1e-4, adjusted via a step schedule. Experiments run on NVIDIA RTX 2080 Ti GPUs, except for MLLM inference, performed on an NVIDIA RTX A6000.

For comparison experiments, we use CD-FSS and MLLM methods as benchmarks to validate the effectiveness of our RoI-Matcher. For CD-FSS, PATNet~\cite{lei2022cross}, RestNet~\cite{huang2023restnet}, DMTNet~\cite{duong2023dmt} and ABCD-FSS~\cite{herzog2024adapt} all use ResNet50~\cite{he2016deep} as the backbone. Because of the difference between task and training dataset, we replace the cross entropy loss of the original model with the Dice loss. For MLLM, we use the open-source method Qwen2-VL-7B~\cite{wang2024qwen2} and InternVL2.5-8B~\cite{chen2024internvl} for comparison. We treat $I_{ref}$, $M_{ref}$, and $I_{tgt}$ as inputs, with the goal of predicting $M_{tgt}$ as the output. Moreover, we employ the in-context learning with a reference-target pair to maintain the formatted output. 
% The prompt is ``You are an excellent assistant proficient in handling multimodal data, particularly images. You 
% will be given two images. The first image comes with a specified region of interest defined by 
% coordinates (x1, y1, x2, y2) where (x1, y1) represents the top-left corner coordinates and (x2, 
% y2) represents the bottom-right corner coordinates of the region within the first image. Your task is to 
% analyze Image 2 and identify the region within it that is related to the region of interest in Image 1. 
% Return the coordinates (in the format of top-left corner coordinates (x1', y1') and bottom-right corner 
% coordinates (x2', y2')) of the related region in Image 2. If you determine that there is no related 
% region in Image 2, clearly state "There is no related region found in Image 2." Please present your response clearly and straightforwardly. ''

\begin{figure}[ht]
\centering
\includegraphics[width=\textwidth]{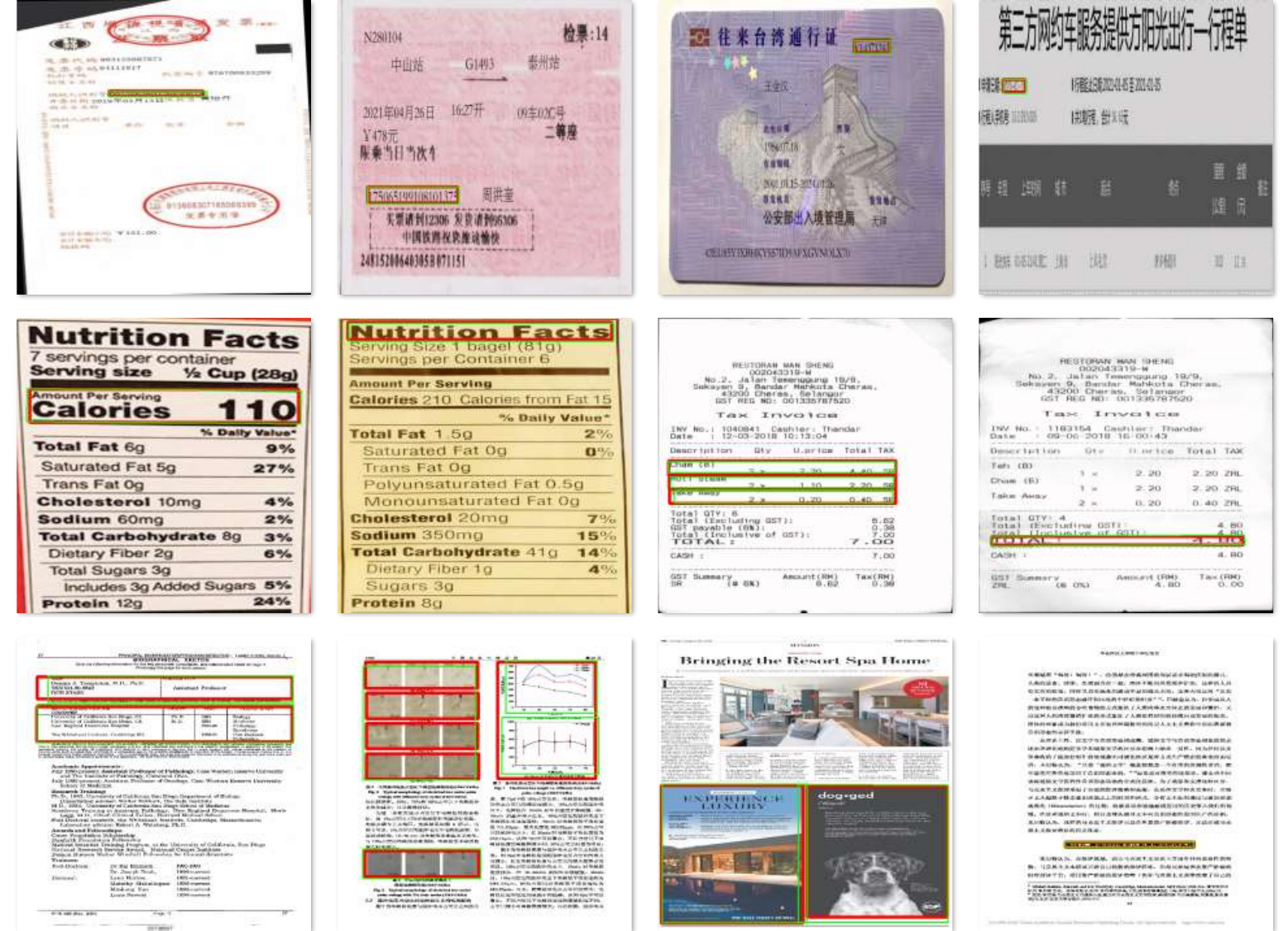}
    \caption{The visualization of RoI-Matcher. The red bounding boxes of each image are ground truths and the green ones are predictions. The difficulty levels I - III are shown from top to bottom in this figure.}
    \label{fig:vis}
    % \vspace{-10pt}
\end{figure}

\subsection{Main Results}

We use representative methods based on CD-FSS and MLLM to compare the RoI-Matching task performance. 
\cref{tab:comparison} shows the results using mIoU and F-Measure, including different datasets and methods. The CD-FSS method performs well in simple scenes, with ResNet achieving 73.2\% mIoU and 63.6\% F-Measure. However, its performance declines significantly in challenging scenes, especially at Level III, dropping to 48.7\% and 15.1\%. While our model also sees some degradation in harder scenarios, it consistently outperforms others across all three levels.
As MLLM, \cref{tab:comparison} shows Qwen2-VL-7B and InternVL2.5-8B achieves comparable performance in Level I and significantly improves over CD-FSS methods in Level II and Level III, especially in F-Measure. However, apart from performing far below our method in the performance of RoI-Matching, the inference speed of Qwen2-VL-7B(3.121s/img) is much slower than our RoI-Matcher (0.056s/img). 
By comparing with methods from other tasks, we verify the necessity of RoI Matching and the superior performance of our RoI Matcher. As shown in \cref{fig:vis}, it achieves strong matching in multi-granular, multi-linguistic and open-set scenarios.

\subsection{Ablation Studies}
% \abl{(space on top of this line)}
\subsubsection{Backbones}
%We load ResNet of different scales on our RoI-Matcher. As shown in \cref{tab:backbone}, RoI-Matcher based on ResNet-50 achieves the best performance, compared with the smaller ResNet-18/34.  We claim that the more powerful backbone is loaded, the better performance our model achieves, because this task relies on deep semantic feature extraction extremely. However, we observe a slow increase in the tendency from ResNet-50 to ResNet-101. we also perform additional experiments to evaluate Swin-T within our RoI-Matcher architecture. Our experiments show that Swin-T underperforms ResNet in RoI-Matcher under identical settings, possibly due to inferior convergence and adaptability on this task. Consequently, we set ResNet-50 as the default setting to balance the performance and efficiency. 
As shown in \cref{tab:backbone},We evaluate different ResNet backbones within RoI-Matcher.  ResNet-50 outperforms smaller variants like ResNet-18 and ResNet-34, as deeper semantic feature extraction is crucial for this task. However, performance gains from ResNet-50 to ResNet-101 are marginal. We also test Swin-T, which underperforms compared to ResNet, likely due to inferior convergence and adaptability on this task. Therefore, we choose the ResNet-50 as the default setting to balance performance and efficiency.

\begin{table}[t]
% \vspace{-0.3cm}
\centering
\caption{Ablation experiments of Backbones.}
\setlength{\tabcolsep}{6pt} 
\label{tab:backbone}
\renewcommand{\arraystretch}{1.0}
\begin{tabular}{lccccc}
\hline
Metrics   & ResNet-18 & ResNet-34 & ResNet-50 & ResNet-101 & Swin-T \\ \hline
mIoU (\%)      &    86.0       &    85.9       &     87.3 &    87.0&    78.3      \\
F-Measure (\%) &     81.0      &    83.0       &    83.9    &   84.2&   50.1   \\ \hline
\end{tabular}
% \vspace{-10pt}
\end{table}

\subsubsection{Training Strategies}

%Due to samples with different difficulty levels in the RoI-Matching-Bench, we compare the joint training strategy with the curriculum training strategy \cite{wang2021survey}. The joint training means three difficulty training subsets participate in the training stage simultaneously. In contrast, the curriculum training is a training strategy that trains a machine learning model from easier data to harder data, which imitates the meaningful learning order in humans
%curricula. \cref{tab:strategy} describes the results of different training strategies. Although curriculum training brings available performance in this task, the joint training achieves a more competitive result on the metrics 
%both mIoU and F-Measure. We argue it is because the difficulty we defined in the benchmark is a human semantic difficulty, rather than a natural difficulty like the image from clear to blurry. Therefore, we use the joint training strategy in the following experiments. 
Due to varying difficulty levels in RoI-Matching-Bench, we compare joint training with curriculum training \cite{wang2021survey}. Joint training uses all three difficulty subsets simultaneously, while curriculum training progresses from easy to hard, mimicking human learning. As shown in \cref{tab:strategy}, although curriculum training yields decent results, joint training performs better in both mIoU and F-Measure.  We argue it is because the difficulty we defined in the benchmark is a human semantic difficulty, rather than a natural difficulty like the image from clear to blurry. Therefore, we use the joint training strategy in the following experiments. 

\begin{table}[t]
% \vspace{-0.3cm}
\centering
\caption{Ablation experiments of training strategies.}
\setlength{\tabcolsep}{6pt}
\label{tab:strategy}
\renewcommand{\arraystretch}{1.0}
\begin{tabular}{lcc}
\hline
Training Strategy   & mIoU (\%) & F-Measure (\%) \\ \hline
Joint Training      &  87.3    &  83.9         \\
Curriculum Training &  79.4    &   83.2        \\ \hline
\end{tabular}
% \vspace{-10pt}
\end{table}
% \vspace{-1cm}

\begin{table}[t]
% \vspace{-0.3cm}
\centering
\caption{Ablation experiments of losses.}
\setlength{\tabcolsep}{6pt}
\label{tab:loss}
\renewcommand{\arraystretch}{1.0}
\begin{tabular}{lcc}
\hline
Setting   & mIoU (\%) & F-Measure (\%) \\ \hline
CrossEntropy + Dice     &  85.4    &  66.8         \\
PANLoss &   85.9 (+0.5\%)   &   83.0 (+16.2\%)        \\ 
\hline
\end{tabular}
% \vspace{-10pt}
\end{table}
% \vspace{-1cm}

\subsubsection{Optimization Target}
%In this subsection, we explore how the setting of loss function influences the performance of RoI-Matcher. The comparison of results is displayed as \cref{tab:loss}. When we use the sum of cross-entropy loss and dice loss as the optimization target, it achieves the performances of 85.4\% of mIoU and 66.8\% of F-Measure. However, the RoI-Matcher with PANLoss increases by 0.5\% and 16.2\%. We analyze why PANLoss achieves higher gains in the F-Measure metric, as it effectively discriminates overlapped regions.
In this subsection, we investigate the impact of different loss functions on RoI-Matcher performance, as shown in \cref{tab:loss}. Using a combination of cross-entropy and dice loss yields 85.4\% mIoU and 66.8\% F-Measure. In contrast, applying PANLoss improves results by 0.5\% in mIoU and 16.2\% in F-Measure. This gain, especially in F-Measure, is attributed to PANLoss's ability to better distinguish overlapping regions.

\begin{table}[t]
% \vspace{-0.3cm}
\centering
\caption{Ablation experiments for efficiency.}   
\setlength{\tabcolsep}{6pt}
\label{tab:eff}
\renewcommand{\arraystretch}{1.0}
\begin{tabular}{lcc}
\hline
Settings   & Param. (M) & Throughput (GFlops) \\ \hline
RoI-Matcher     &   34.5   &      149.3     \\
FPNC $\rightarrow$ FPN &    43.6 (+26.3\%)  &     232.8 (+55.9\%)     \\  
wo Grid Sampling &   34.5 (0\%)   &    319.5 (+114\%)      \\ 
\hline
\end{tabular}
\end{table}
% \vspace{-1cm}

\begin{table}[t]
\centering
\caption{Comparison between Pre-mask and Post-mask.}
\setlength{\tabcolsep}{6pt}
\label{tab:post}
\renewcommand{\arraystretch}{1.0}
\begin{tabular}{lcc}
\hline
 Settings   & mIoU (\%) & F-Measure (\%) \\ \hline
Post-mask  &     87.1     &         68.4    \\
Pre-mask   &    87.3 (+0.2\%)     &        83.9 (+22.7\%)      \\
\hline
\end{tabular}
\vspace{-10pt}
\end{table}

% \begin{table}[htbp]
% % \vspace{-0.3cm}
% \centering
% \caption{Comparison of Seg. heads.}
% \setlength{\tabcolsep}{6pt}
% \label{tab:head}
% \renewcommand{\arraystretch}{1.0}
% \begin{tabular}{lccc}
% \hline
% Seg. Head   & mIoU (\%) & F-Measure (\%) \\
% \hline
% Mask R-CNN      &    85.8      &      75.1      \\
% FCN      &    87.3 (+1.7\%)       &      83.9 (+11.7\%)      \\
% \hline
% \end{tabular}
% \end{table}
% \vspace{-1cm}

\subsubsection{Efficiency} %In our baseline, we introduce two efficient modules. We conduct ablations to validate the efficiency of our baseline as shown in \cref{tab:eff}. The first line of \cref{tab:eff} shows the parameter number and throughput of our baseline, which are 4.5 M and 149.3 GFlops respectively. However, when we replace the classical FPNC with Feature Pyramid Network (FPN)  used in our network, the parameter and the throughput increase by 26.3\% and 55.9\% substantially. Furthermore, we test the situation without grid sampling. Compared with our baseline, the parameter remains stable but the throughput increases by 114\%, which slows down the inference speed significantly. The two ablations guarantee that these two micro designs are efficient.
In our baseline, we introduce two efficient modules and conduct ablations to validate their effectiveness, as shown in \cref{tab:eff}. The first row reports 4.5M parameters and 149.3 GFlops. Replacing our FPNC with Feature Pyramid Network (FPN) increases
parameters by 26.3\% and throughput by 55.9\%. Removing grid sampling keeps the parameter count unchanged but increases throughput by 114\%, significantly reducing inference time. These ablations confirm the efficiency of both design choices.

\subsubsection{Pre-mask vs Post-mask}We conduct ablation studies to examine the optimal placement of the mask operation relative to our FPNC. As presented in \cref{tab:post}, applying the mask after FPNC (Post-mask) can result in notable information loss in the source domain, indicating that early masking is more effective for preserving semantic features.

% \subsubsection{Segmentation head}We perform ablations on different segmentation heads, as shown in \Cref{tab:head}. The experimental results show that the FCN segmentation head achieves an mIoU that is 1.7\% higher and an F-measure that is 11.7\% higher than those of Mask R-CNN. Given that there are a large number of fine-grained matches in our dataset, the FPN structure may be more effective at capturing detailed and boundary information, allowing it to retain such features more accurately and achieve better results.

\section{Limitations}

%Although we propose a novel task RoI-Matching and design a feasible baseline, several limitations still need to be solved. First, we have only developed a 1-shot RoI-Matcher, which means that the model is unable to self-adjust and improve the matching results through multiple visual prompts. Secondly, our baseline is a uni-modal matching model that does not incorporate any linguistic information. We will discuss the integration of a cross-modal design in our future work.
Although we propose a novel task RoI-Matching and design a feasible baseline, several limitations remain. First, our current RoI-Matcher is a 1-shot model, unable to self-adjust or refine matching results through multiple visual prompts. Secondly, it is a uni-modal model that does not incorporate any linguistic information. Furthermore, We have not yet attempted repeated inference across multiple RoIs, but we propose a feasible strategy for optimization. When given a multi-channel source mask, RoI-Matcher can generate corresponding masks with the same number of channels for parallel inference. We plan to address these limitations in future work.

\section{Conclusion}

In this paper, we propose the class-agnostic region-of-interest matching task creatively, to obtain the target corresponding regions from the reference target and customized visual prompt, in a flexible, efficient, multi-granularity, and open-set manner. Then we construct an abundant benchmark RoI-Matching-Bench with three
levels of difficulties, which allows researchers to evaluate the performance of RoI-Matching approach under real-world conditions. We also develop an effective and efficient baseline RoI-Matcher. The extensive experiments show that our simple method outperforms other CD-FSS and MLLM methods. Several ablations also indicate our subtle designs are effective and efficient. In the future, we hope to enhance our RoI-Matcher in the perspective of multi-modal and multi-interactive.

\section*{Acknowledgments}
Supported by the National Natural Science Foundation of China (Grant NO 62376266 and 62406318), Key Laboratory of Ethnic Language Intelligent Analysis and Security Governance of MOE, Minzu University of China, Beijing, China.

% \clearpage

\bibliographystyle{unsrt}
\bibliography{sample-base}
\end{document}